\pdfoutput=1

\documentclass[11pt]{article}

\usepackage[preprint]{acl}

\usepackage{times}
\usepackage{latexsym}

\usepackage[T1]{fontenc}

\usepackage[utf8]{inputenc}

\usepackage{microtype}

\usepackage{inconsolata}

\usepackage{xcolor} 

\usepackage{amsmath}
\usepackage{bm}
\usepackage{booktabs}
\usepackage{caption}
\usepackage{makecell}
\usepackage{multirow}
\usepackage{wrapfig}
\usepackage{listings}
\usepackage{graphicx}
\usepackage{xcolor}

\definecolor{maroon}{cmyk}{0, 0.87, 0.68, 0.32}
\definecolor{halfgray}{gray}{0.55}
\definecolor{ipython_frame}{RGB}{207, 207, 207}
\definecolor{ipython_bg}{RGB}{247, 247, 247}
\definecolor{ipython_red}{RGB}{186, 33, 33}
\definecolor{ipython_green}{RGB}{0, 128, 0}
\definecolor{ipython_cyan}{RGB}{64, 128, 128}
\definecolor{ipython_purple}{RGB}{170, 34, 255}

\usepackage{listings}
\lstdefinelanguage{iPython}{
    morekeywords={access,and,break,class,continue,def,del,elif,else,except,exec,finally,for,from,global,if,import,in,is,lambda,not,or,pass,print,raise,return,try,while},%
    morekeywords=[2]{abs,all,any,basestring,bin,bool,bytearray,callable,chr,classmethod,cmp,compile,complex,delattr,dict,dir,divmod,enumerate,eval,execfile,file,filter,float,format,frozenset,getattr,globals,hasattr,hash,help,hex,id,input,int,isinstance,issubclass,iter,len,list,locals,long,map,max,memoryview,min,next,object,oct,open,ord,pow,property,range,raw_input,reduce,reload,repr,reversed,round,set,setattr,slice,sorted,staticmethod,str,sum,super,tuple,type,unichr,unicode,vars,xrange,zip,apply,buffer,coerce,intern},%
    sensitive=true,%
    morecomment=[l]\#,%
    morestring=[b]',%
    morestring=[b]",%
    morestring=[s]{'''}{'''},
    morestring=[s]{"""}{"""},
    morestring=[s]{r'}{'},
    morestring=[s]{r"}{"},%
    morestring=[s]{r'''}{'''},%
    morestring=[s]{r"""}{"""},%
    morestring=[s]{u'}{'},
    morestring=[s]{u"}{"},%
    morestring=[s]{u'''}{'''},%
    morestring=[s]{u"""}{"""},%
    literate=
    {á}{{\'a}}1 {é}{{\'e}}1 {í}{{\'i}}1 {ó}{{\'o}}1 {ú}{{\'u}}1
    {Á}{{\'A}}1 {É}{{\'E}}1 {Í}{{\'I}}1 {Ó}{{\'O}}1 {Ú}{{\'U}}1
    {à}{{\`a}}1 {è}{{\`e}}1 {ì}{{\`i}}1 {ò}{{\`o}}1 {ù}{{\`u}}1
    {À}{{\`A}}1 {È}{{\'E}}1 {Ì}{{\`I}}1 {Ò}{{\`O}}1 {Ù}{{\`U}}1
    {ä}{{\"a}}1 {ë}{{\"e}}1 {ï}{{\"i}}1 {ö}{{\"o}}1 {ü}{{\"u}}1
    {Ä}{{\"A}}1 {Ë}{{\"E}}1 {Ï}{{\"I}}1 {Ö}{{\"O}}1 {Ü}{{\"U}}1
    {â}{{\^a}}1 {ê}{{\^e}}1 {î}{{\^i}}1 {ô}{{\^o}}1 {û}{{\^u}}1
    {Â}{{\^A}}1 {Ê}{{\^E}}1 {Î}{{\^I}}1 {Ô}{{\^O}}1 {Û}{{\^U}}1
    {œ}{{\oe}}1 {Œ}{{\OE}}1 {æ}{{\ae}}1 {Æ}{{\AE}}1 {ß}{{\ss}}1
    {ç}{{\c c}}1 {Ç}{{\c C}}1 {ø}{{\o}}1 {å}{{\r a}}1 {Å}{{\r A}}1
    {€}{{\EUR}}1 {£}{{\pounds}}1
    {^}{{{\color{ipython_purple}\^{}}}}1
    {=}{{{\color{ipython_purple}=}}}1
    {+}{{{\color{ipython_purple}+}}}1
    {*}{{{\color{ipython_purple}$^\ast$}}}1
    {/}{{{\color{ipython_purple}/}}}1
    {+=}{{{+=}}}1
    {-=}{{{-=}}}1
    {*=}{{{$^\ast$=}}}1
    {/=}{{{/=}}}1,
    literate=
    *{-}{{{\color{ipython_purple}-}}}1
     {?}{{{\color{ipython_purple}?}}}1,
    identifierstyle=\color{black}\ttfamily,
    commentstyle=\color{ipython_cyan}\ttfamily,
    stringstyle=\color{ipython_red}\ttfamily,
    keepspaces=true,
    showspaces=false,
    showstringspaces=false,
    rulecolor=\color{ipython_frame},
    frame=single,
    frameround={t}{t}{t}{t},
    framexleftmargin=0mm,
    numberstyle=\tiny\color{halfgray},
    backgroundcolor=\color{ipython_bg},
    basicstyle=\footnotesize\ttfamily,
    keywordstyle=\color{ipython_green}\ttfamily,
    aboveskip=1.2em,
    belowskip=1.2em,
}

\usepackage{amsmath,amsfonts}
\usepackage[utf8]{inputenc} 
\usepackage[T1]{fontenc}    
\usepackage{url}            
\usepackage{booktabs}       
\usepackage{amsfonts}       
\usepackage{nicefrac}       
\usepackage{microtype}      
\usepackage{xcolor}         
\usepackage{algorithm}
\usepackage{algorithmic}
\usepackage{float}         
\usepackage[skins]{tcolorbox} 

\usepackage{enumitem}

\usepackage{pifont}
\newlength\savewidth

\makeatletter\renewcommand\paragraph{\@startsection{paragraph}{4}{\z@}
  {.5em \@plus1ex \@minus.2ex}{-.5em}{\normalfont\normalsize\bfseries}}\makeatother
\usepackage{xspace}

\newcommand{\ModelName}{LMFlow}

\tcbuselibrary{breakable} 
\newtcolorbox[auto counter, number within=section, list type=subsubsection, list inside=toc]{sectionbox}[2][]{
colback=white!98!gray, colframe=black, 
colbacktitle=white!90!gray, coltitle=black, 
fonttitle=\bfseries,
title={#2}, 
list entry={Comment \thetcbcounter\quad}
}

\newcommand{\hmm}{\textbf{Human: }}     
\newcommand{\assi}{\textbf{Assistant: }}     
\newcommand{\araft}{\textbf{LLaMA-RAFT: }}     
\newcommand{\allama}{\textbf{LLaMA: }}     
\newcommand{\asft}{\textbf{LLaMA-SFT: }}     
\newcommand{\appo}{\textbf{LLaMA-PPO: }}

\usepackage{xcolor}

\lstdefinestyle{lfonts}{
basicstyle = \footnotesize\ttfamily,
stringstyle = \color{purple},
keywordstyle = \color{blue!60!black}\bfseries,
commentstyle = \color{olive}\scshape,
}

\lstdefinestyle{lnumbers}{
numbers = left,
numberstyle = \tiny,
numbersep = 1em,
firstnumber = 1,
stepnumber = 1,
}

\lstdefinestyle{llayout}{
breaklines = true,
tabsize = 2,
columns = flexible,
}

\lstdefinestyle{lgeometry}{
xleftmargin = 20pt,
xrightmargin = 0pt,
frame = tb,
framesep = \fboxsep,
framexleftmargin = 20pt,
}

\lstdefinestyle{lgeneral}{
style = lfonts,
style = lnumbers,
style = llayout,
style = lgeometry,
}

\lstdefinestyle{python}{
language = {Python},
style = lgeneral,
}

\usepackage{lastpage}

\title{\raisebox{0mm}{\includegraphics[trim=0 0 -20 0, clip, width=0.8cm]{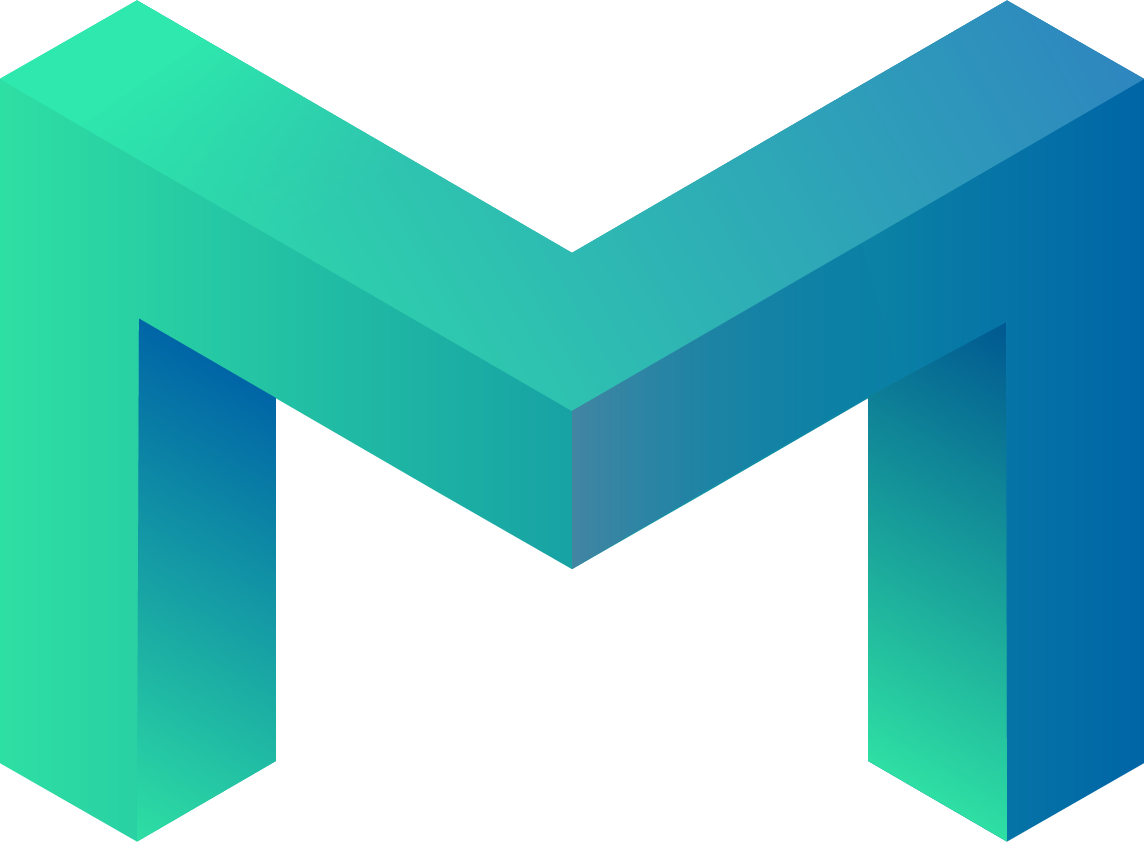}} LMFlow: An Extensible Toolkit for Finetuning and Inference \\ of Large Foundation Models}

\author{Shizhe Diao$^{\heartsuit*}$, ~ Rui Pan$^{\heartsuit*}$, ~ Hanze Dong$^{\heartsuit*}$, \\ ~ \bf KaShun Shum$^{\heartsuit}$, ~ \bf Jipeng Zhang$^{\heartsuit}$, ~ \bf Wei Xiong$^{\spadesuit}$, ~ \bf Tong Zhang$^{\spadesuit}$\\
  $^{\heartsuit}$The Hong Kong University of Science and Technology\\
  $^{\spadesuit}$University of Illinois Urbana-Champaign\\
  \texttt{\{sdiaoaa, rpan, hdongaj\}@ust.hk}\\
  \texttt{tozhang@illinois.edu}\\
  \\
}

\begin{document}
\maketitle
\def\thefootnote{*}\footnotetext{Equal Contribution.}
\def\thefootnote{\arabic{footnote}}
\begin{abstract}
Foundation models have demonstrated a great ability to achieve general human-level intelligence far beyond traditional approaches. 
As the technique keeps attracting attention from the AI community, an increasing number of foundation models are becoming publicly accessible.
However, a significant shortcoming of most of these models lies in their performance in specialized-domain and task-specific applications, necessitating domain- and task-aware fine-tuning to develop effective scientific language models.
As the number of available foundation models and specialized tasks keeps growing, the job of training scientific language models becomes highly nontrivial. 
In this paper, we initiate steps to tackle this issue.
We introduce an extensible and lightweight toolkit, {\ModelName}, which aims to simplify the domain- and task-aware finetuning of general foundation models.
LMFlow offers a complete finetuning workflow for a foundation model to support specialized training with limited computing resources.
Furthermore, it supports continuous pretraining, instruction tuning, parameter-efficient finetuning, alignment tuning, inference acceleration, long context generalization, model customization, and even multimodal finetuning, along with carefully designed and extensible APIs. 
This toolkit has been thoroughly tested and is available at \url{https://github.com/OptimalScale/LMFlow}.\footnote{Video demonstrations trained and deployed by LMFlow:
\begin{itemize}
    \item Emotional Companion bot: \url{https://www.youtube.com/watch?v=BDSME4f2AjU}
    \item Multimodal reasoning-based object detection bot: \url{https://www.youtube.com/watch?v=YXNyh6bGqyI}
\end{itemize}
}
\end{abstract}

\section{Introduction}

\begin{table*}[t]
\centering
\def\w{20pt}
  \setlength{\tabcolsep}{1.8mm}{
\begin{tabular}{l|cccccccccc}
Packages & Cont. PT & FT & RLHF & Deploy. & Adapt. & Acc. & LC & VE & MM \\
\midrule
Transformers~\citep{wolf-etal-2020-transformers} & \ding{51} & \ding{51} &  &  & & & & \ding{51} & \ding{51} \\
Accelerate~\citep{accelerate} & \ding{51} & \ding{51} & & & & \ding{51} & \\
Deepspeed~\citep{rasley2020deepspeed} & \ding{51} & \ding{51} & & &  & \ding{51} \\
Trl~\citep{vonwerra2022trl} & & & \ding{51} & & \\
{\ModelName} (ours) & \ding{51} & \ding{51} & \ding{51} & \ding{51} & \ding{51} & \ding{51} & \ding{51} & \ding{51} & \ding{51} \\
  \end{tabular}
  }
\caption{
Comparison with competing packages.
\textbf{Cont. PT}: continuous pretraining.
\textbf{FT}: finetuning.
\textbf{RLHF}: reinforcement learning from human feedback.
\textbf{Deploy.}: deployment.
\textbf{Adapt.}: domain/task adaptation.
\textbf{Acc.}: acceleration techniques for finetuning and inference.
\textbf{LC}: long context generalization.
\textbf{VE}: vocabulary extension.
\textbf{MM}: multimodal training.
}
\label{tab:tuning_glossary}
\end{table*}

\begin{figure*}[t]
    \centering
    \includegraphics[scale=0.45, trim=0 80 0 0, clip]{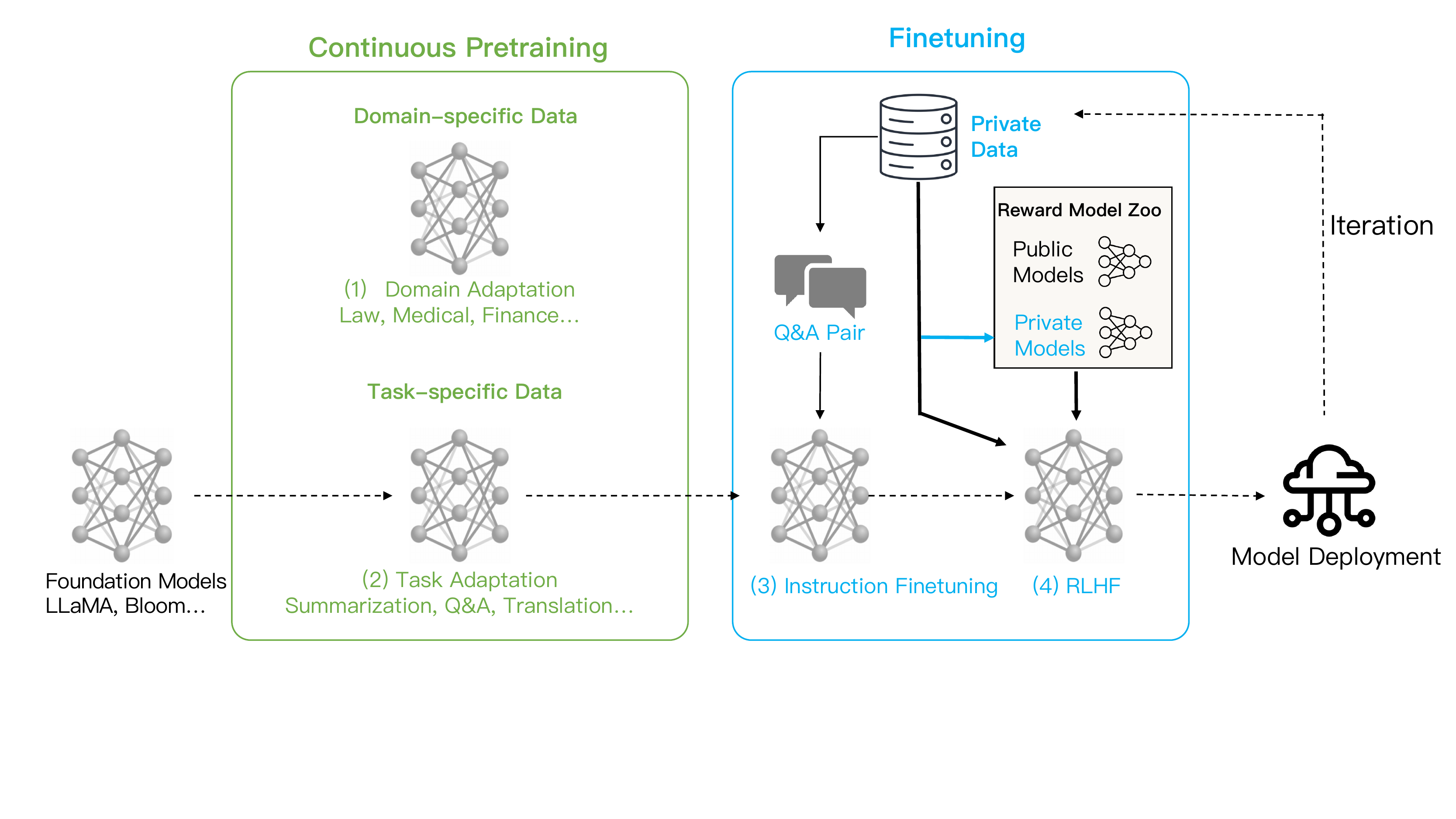}
    \vskip -1em
    \caption{The system design of LMFlow. 
    Starting from a publicly available foundation model, there are four possible stages including (1) domain adaptation, (2) task adaptation, (3) instruction finetuning, and (4) reinforcement learning with human feedback.
    }
    \label{fig:model_arch}
\end{figure*}

Foundation models (FMs), and in particular large language models (LLMs), have demonstrated general abilities to perform different tasks beyond what was possible previously. 
While a number of pretrained large models, including GPT-J~\citep{gpt-j}, Bloom~\citep{scao2022bloom}, LLaMA~\citep{touvron2023llama, touvron2023llama2}, etc., are publicly available and have already been incorporated into the Hugging Face model repository \citep{huggingface}, there is no publicly available toolkit that can be easily used to perform finetuning and inference for these different models. 
For specialized domains or tasks, it is necessary to further finetune such LLMs to achieve improved performance on such domains or tasks. 
The purpose of this package is to offer a simple-to-use and lightweight toolkit so that developers and researchers can perform efficient finetuning and inference of scientific language models with limited resources. 
The typical processes to train a scientific language model are shown in~Figure~\ref{fig:model_arch}, which include: 
\begin{itemize}
[leftmargin=*,label=$\bullet$,noitemsep,partopsep=0pt,topsep=0pt,parsep=0pt]
    \item Continuous pretraining on datasets in special domains and tasks so that a foundation model can acquire domain- and task-specific knowledge.
    It normally contains domain or task adaptation.
    \item Instruction tuning to teach a foundation model the capability to follow these specialized natural language instructions and perform tasks required by such instructions. 
    \item Reinforcement learning from human feedback (RLHF) to align a foundation model to human preference (for example, helpfulness, harmlessness, and honesty). 
\end{itemize}

{\ModelName} enhances and streamlines the aforementioned fine-tuning procedures, enabling the efficient and effective training of a scientific language model.
We focus on improving training speed.
For example, it only takes one Nvidia 3090 GPU and five hours to train a medical LLaMA comparable to ChatGPT, based on a 7-billion-parameter LLaMA model.
In addition to speed, we also aspire to achieve higher model performance.
We used this framework to train medical LLaMA, a series of models with 7-billion, 13-billion, 33-billion, and 65-billion parameters, on a single machine and have released the model weights for academic research. 
Using LMFlow, anyone can train their own scientific or personalized language models.
Each person can choose the appropriate foundation model according to their available resources, for tasks such as question answering, companionship, and expert consultations in various domains.
The larger the model and data size, the longer the training time and the better the results. 
Compared with existing packages, {\ModelName} encompasses a multitude of features that are absent in others, such as the support for long context generalization, as shown in Table~\ref{tab:tuning_glossary}. 
Most importantly, {\ModelName} stands out as a comprehensive, full-cycle foundation model adaptation toolkit. 
While other packages excel in specific areas like finetuning, they lack functionalities like RLHF and others. 
To our knowledge, {\ModelName} is the first to offer a complete pipeline that integrates all these processes. 
This holistic toolkit allows for more robust and adaptable language model training and inference, setting a new standard in the field of natural language processing.

\section{Related Work}

In recent years, the finetuning of large language models (LLMs) has gained significant attention, especially for scientific domain applications. 
The necessity of adapting these general-purpose models to specific domains or tasks  has led to the development of various scientific language models.
\citet{lehman2023we} conducted an extensive empirical analysis on the performance of various language models in clinical tasks and found that specialized clinical models, even smaller in size, significantly outperform larger general-domain models when finetuned on domain-specific data. 
This emphasizes the importance of domain specialization in achieving higher accuracy in safety-critical fields like healthcare.
Therefore, a series of scientific large models have emerged, including but not limited to:
language models for Science~\citep{beltagy2019scibert, luu2021explaining, taylor2022galactica}, Mathematics~\citep{yue2023mammoth, yu2023metamath, gao2023g}, Physics~\citep{nguyen2023astrollama, zheng2023marinegpt, perkowski2024astrollama}, Chemistry and Materials Science~\citep{cao2023instructmol, shetty2023general, rubungo2023llm}, Biology and Medicine~\citep{lee2020biobert, zhang2023huatuogpt, singhal2023towards, wu2023pmc, han2023medalpaca, wang2023huatuo, yang2024pllama}, and Information Retrieval~\citep{lassance2023experimental}
We recommend readers to refer to a paper list of scientific language models~\footnote{\url{https://github.com/yuzhimanhua/Awesome-Scientific-Language-Models}}, which includes a more comprehensive range of works related to scientific language models.
Among these works, LMFlow has successfully helped in training AstroLLaMA-Chat~\citep{perkowski2024astrollama} and MarineGPT~\citep{zheng2023marinegpt}. 
The Medical LLaMA trained in the medical domain within this paper also demonstrates the effectiveness of LMFlow.
In summary, our proposed LMFlow offers a comprehensive toolkit for efficient and effective finetuning of foundation models across various specialized domains.

\section{Toolkit Overview}

\subsection{System Design}
An illustration of the LMFlow system design is shown in Figure~\ref{fig:model_arch}.
There are four stages for improving the performance of a publicly available foundation model.
The first stage is domain adaptation, which involves modifying the model to better handle a specific domain by training the model on that domain. 
The second stage is task adaptation, which involves adapting the model to perform a specific task, such as summarization, question-answering, and translation. 
The third stage is instruction finetuning, which involves adjusting the model's parameters based on instructional question-answer pairs. 
The final stage is reinforcement learning with human feedback, which involves using human feedback to further align the model to human preference.
{\ModelName} provides a complete finetuning workflow for these four stages, supporting large language models' specialized training with limited computing resources.
Especially, {\ModelName} supports the following key features:
\begin{itemize}
[leftmargin=*,label=$\bullet$,noitemsep,partopsep=0pt,topsep=0pt,parsep=0pt]
    \item Finetuning Acceleration and Memory Optimization: LoRA~\citep{hulora}, FlashAttention~\citep{dao2022flashattention, dao2023flashattention}, Gradient Checkpointing, and Deepspeed Zero3.
    \item Inference Acceleration: Speculative Decoding~\citep{leviathan2023fast}, LLaMA Inference on CPU, and FlashAttention~\citep{dao2022flashattention, dao2023flashattention}.
    \item Alignment Tuning: An implementation of our proposed novel alignment algorithm RAFT~\citep{dong2023raft} (Reward rAnked FineTuning) to simply RLHF pipeline for generative models. 
    \item Long Context Generalization: Position Interpolation for LLaMA~\citep{chen2023extending}.
    \item Model Customization: Vocabulary Extension.
    \item Multimodal: Finetuning Multimodal Chatbot for reasoning-based object detection~\citep{pi2023detgpt}.
\end{itemize}

\subsection{Installation}
LMFlow has been fully tested on Linux OS (Ubuntu 20.04) and can be installed by executing the following commands.
\begin{lstlisting}[language=iPython]
$ git clone https://github.com/OptimalScale/LMFlow.git
$ cd LMFlow
$ conda create -n lmflow python=3.9 -y
$ conda activate lmflow
$ pip install -e .
\end{lstlisting}

\subsection{Data Format}
LMFlow accepts several \texttt{.json} files as input.
Users can provide a list of \texttt{.json} files under a specified dataset directory. 
For example,
\begin{lstlisting}[style = python]
|- path_to_dataset
  |- data_1.json
  |- data_2.json
  |- another_data.json
  |- ...
\end{lstlisting}
Each json file shall have the following format (three instances with four keys for example),
\begin{lstlisting}[style = python]
{
  "type": "TYPE",
  "instances": [
    {
        "KEY_1": "VALUE_1.1",
        "KEY_2": "VALUE_1.2",
        "KEY_3": "VALUE_1.3",
        "KEY_4": "VALUE_1.4",
    },
    {
        "KEY_1": "VALUE_2.1",
        "KEY_2": "VALUE_2.2",
        "KEY_3": "VALUE_2.3",
        "KEY_4": "VALUE_2.4",
    },
    {
        "KEY_1": "VALUE_3.1",
        "KEY_2": "VALUE_3.2",
        "KEY_3": "VALUE_3.3",
        "KEY_4": "VALUE_3.4",
    },
  ]
}
\end{lstlisting}

where the \texttt{TYPE} indicates the dataset type and defines the set of keys \texttt{\{ KEY\_1, KEY\_2, ... \}} and their corresponding interpretations. 
Two supported \texttt{.json} formats are detailed as follows.

\paragraph{TextOnly}
This is the most common dataset type, which only contains raw texts in each sample. This type of dataset can be used as the training set for text decoder models, or the input of decoder models / encoder-decoder models. Its format is as follows (three instances, for example),

\begin{lstlisting}[style = python]
{
  "type": "text_only",
  "instances": [
    {  "text": "SAMPLE_TEXT_1" },
    {  "text": "SAMPLE_TEXT_2" },
    {  "text": "SAMPLE_TEXT_3" },
  ]
}
\end{lstlisting}

\paragraph{Text2Text}
This is the dataset type mostly used for inferencing, which contains a pair of texts in each sample. 
This type of dataset can be used as the training set for text encoder-decoder models, or question-answer pairs for evaluating model inferences. Its format is as follows (three instances for example),
\begin{lstlisting}[style = python]
{
  "type": "text2text",
  "instances": [
    {
        "input": "SAMPLE_INPUT_1",
        "output": "SAMPLE_OUTPUT_1",
    },
    {
        "input": "SAMPLE_INPUT_2",
        "output": "SAMPLE_OUTPUT_2",
    },
    {
        "input": "SAMPLE_INPUT_3",
        "output": "SAMPLE_OUTPUT_3",
    },
  ]
}
\end{lstlisting}

\subsection{Continuous Pretraining}
The endeavor to bridge the divide between pretraining domains and downstream domains has led to the adoption of a prevalent approach, known as continuous pretraining~\citep{beltagy2019scibert, alsentzer2019publicly, huang2019clinicalbert, lee2020biobert}, which involves the ongoing pretraining on an extensive collection of unlabeled data that is specific to a given domain.
LMFlow supports continuous pretraining natively, which is an effective way to adapt LLMs to a specific domain.
Users just need to collect a set of unlabeled data and prepare them to \texttt{TextOnly} data format. 
The following process will be handled by autoregressive training.

\subsection{Instruction Tuning}
Instruction tuning~\citep{sanhmultitask, weifinetuned, chung2022scaling, muennighoff2022crosslingual, wang2022self}, also called supervised finetuning, is an approach to enhance the performance of language models by training them to follow natural language instructions. 
This involves training the model on a small set of task-specific data, most of which are in prompt-answer format, including positive or negative examples, prompts, constraints, and other elements commonly present in human language. 
Instruction tuning enables LLMs to provide more accurate and relevant responses to user queries, making them more effective conversational agents.

\begin{table*}[t]
\centering
\footnotesize
\begin{tabular}{l|c|c|c|c|c|c|c}
\toprule
\textsc{Model} & anatomy & \begin{tabular}[c]{@{}c@{}}clinical\\ knowledge\end{tabular}    & \begin{tabular}[c]{@{}c@{}}college\\ biology\end{tabular}   & \begin{tabular}[c]{@{}c@{}}college\\ medicine\end{tabular}  & \begin{tabular}[c]{@{}c@{}}medical\\ genetics\end{tabular}   & \begin{tabular}[c]{@{}c@{}}professional\\ medicine\end{tabular}   & Average \\
\midrule
LLaMA 33B & 39.2 & 40.3 & 44.4 & 32.9 & 36.0 & 43.0 & 39.3 \\
Galactica 30B & 32.5 & 26.0 & 30.5 & 25.4 & 39.0 & 23.1 & 29.4 \\
Galactica 120B & \textbf{58.5} & 59.2 & 68.7 & 57.2 & 68.0 & 59.6 & 61.9 \\
OPT 175B & 28.9 & 21.9 & 30.6 & - & 35.0 & 27.9 &  - \\
BLOOM 176B & 37.0 & 29.8 & 28.5 & - & 36.0 & 25.4 & - \\
Gopher 280B & 56.3 & 67.2 & 70.8 & 60.1 & 69.0 & 64.0 & 64.6\\
GPT3.5 & 56.3 & \textbf{69.8} & \textbf{72.2} & \textbf{61.3} & \textbf{70.0} & \textbf{70.2} & \textbf{66.6} \\
\midrule
Task-tuned LLaMA 33B (LoRA) & 51.8 & 65.2 & 70.1 & 58.3 & 65.6 & 66.5 & 62.9 \\
\bottomrule
\end{tabular}
\caption{The performance on Massive Multitask Language Understanding (MMLU) benchmark.
\textbf{Bold} represents the best among each dataset.
}
\label{tab:mmlu}
\end{table*}

\begin{table*}[t]
\centering
\footnotesize
\begin{tabular}{l|c|c|c|c}
\toprule
\textsc{Model} & PubMedQA (ID) & MedQA-USMLE (OOD) & MedMCQA (ID) & Average \\
\midrule
Human (pass) & - & 60.0 & 50.0 & - \\
Human (expert) & 78.0 & 87.0 & 90.0 & 85.0 \\
\midrule
InstructGPT-175B & 73.2 & 46.0 & 44.0 & 54.4 \\
ChatGPT & 63.9 & \textbf{57.0} & 44.7 & 55.2 \\
LLaMA-7B & 5.2 & 27.1 & 24.3 & 18.9 \\
LLaMA-33B & 1.8 & 43.4 & 30.3 & 25.2 \\
\midrule
Task-tuned LLaMA-7B (full) & \textbf{75.1} & 44.5 & 49.9 & 56.5 \\
Task-tuned LLaMA-33B (LoRA) & 74.0 & 51.3 & \textbf{50.2} & \textbf{58.5} \\
\bottomrule
\end{tabular}
\caption{
The overall performance of task-tuned LLaMA models and the comparison with human and existing models on three medical datasets.
PubMedQA and MedMCQA are evaluated on in-domain tests and MedQA-USMLE is evaluated on the out-of-domain test.
\textbf{Bold} represents the best among each dataset.
}
\label{tab:medical_qa}
\end{table*}

\subsection{RLHF as Finetuning}
There is a growing need to explore alternative pretraining objectives that can guide LLMs to generate text that aligns with human preferences.
By doing so, we can ensure that LLMs produce text that is more helpful, honest, and harmless for humans, which are called `HHH' rules~\citep{askell2021general}. \citet{ouyang2022training} divides the alignment process into three steps, including SFT, reward modeling, and RLHF (reward optimization). 
We have integrated all of these steps into our LMFlow framework. 
For reward optimization, PPO has been shown to be effective in various studies~\citep{schulman2017proximal, engstrom2020implementation}. 
However, it relies on a trial-and-error approach through interaction with the environment, making it less stable and efficient than supervised learning~\citep{choshen2019weaknesses}. 
To address this, we propose and implement a new alignment method for generative models called RAFT~\citep{dong2023raft}. 
RAFT utilizes a reward model to rank the output of the generative model, allowing us to continue training using supervised finetuning (SFT)-like techniques with the selected samples. 
This approach encourages the generative model to prioritize samples with higher rewards and offers significant computational advantages over PPO, resulting in substantial savings in memory and gradient computations. 
Moreover, due to the stability of SFT-like training, our approach demonstrates lower sample complexity and requires fewer learnable parameters, making it easily adaptable to any generative model. 
We believe our novel alignment algorithm represents a competitive and innovative approach that contributes to the well-behaved behavior of generative models.

\subsection{Efficient Tuning}
LMFlow supports low-rank adaptation (LoRA)~\citep{hulora} tuning based on the implementation of \texttt{huggingface/peft}~\citep{peft}~\footnote{\url{https://github.com/huggingface/peft}}.
LoRA is an efficient tuning method that involves freezing the weights of the pretrained model and incorporating trainable rank decomposition matrices into each layer of the Transformer architecture.
This approach significantly reduces the number of trainable parameters. 
On top of that, LMFlow integrates the feature of QLoRA~\citep{dettmers2023qlora}, allowing the training of even larger-sized LLMs.

\subsection{Inference}
LMFlow developed an easy-to-use inference interface for LLMs, which
supports parameter partitioning with zero-offload strategies as introduced by Deepspeed~\citep{ren2021zero}.
In LMFlow, the inference interface is provided by an \texttt{inferencer} class.
The \texttt{inferencer} contains two important inference classes: \texttt{inference} and \texttt{stream\_inference}.
The distinction lies in whether the output is printed word by word in real-time. Speculative decoding is further supported in \texttt{SpeculativeInferencer}.

\begin{table*}[t]
\centering
\footnotesize
\begin{tabular}{l|c|c|c|c|c}
\toprule
\textsc{Model}  & ARC-C  & HellaSwag & MMLU & TruthfulQA  & Average   \\
 \midrule
 \multicolumn{6}{c}{\textit{7B}}
  \\ \midrule
LLaMA-7B~\citep{touvron2023llama} & 46.6 & 75.6 & 34.2 & 34.1 & 47.6 \\
Baize-7B-v2~\citep{xu2023baize}  & 44.5 & 73.3 & 35.6 & 40.8 & 48.6 \\
MPT-7B~\citep{MosaicML2023Introducing}  & 47.7 & 77.7 & 35.6 & 33.4 & 48.6 \\
Falcon-7B~\citep{refinedweb}  & 47.9 & 78.1 & 35.0 & 34.3 & 48.8 \\
Robin-7B-v2 & 49.4 & 74.6 & 39.8 & 43.0 & 51.7 \\
 \midrule
 \multicolumn{6}{c}{\textit{13B}}
  \\ \midrule
Alpaca-13B~\citep{alpaca}  & 51.9 & 77.6 & 37.6 & 39.6 & 51.7 \\
LLaMA-13B~\citep{touvron2023llama}  & 50.8 & 78.9 & 37.7 & 39.9 & 51.8 \\
Vicuna-13B~\citep{zheng2023judging}  & 47.4 & 75.2 & 39.6 & 49.8 & 53.7 \\
Baize-13B-v2~\citep{xu2023baize}  & 50.3 & 77.1 & 39.4 & 48.3 & 53.8 \\
Robin-13B-v2 & 56.5 & 80.4 & 48.8 & 50.8 & 59.1  \\
 \midrule
 \multicolumn{6}{c}{\textit{>30B}}
  \\ \midrule
LLaMA-33B~\citep{touvron2023llama}  & 57.1 & 82.6 & 45.7 & 42.3 & 56.9 \\
LLaMA-65B~\citep{touvron2023llama}  & 57.8 & 84.2 & 48.8 & 42.3 & 58.3 \\
Falcon-40B~\citep{refinedweb}  & 61.9 & \textbf{85.3} & 52.7 & 41.7 & 60.4 \\
Guanaco-65B-merged~\citep{dettmers2023qlora}  & 60.2 & 84.6 & 52.7 & 51.3 & 62.2 \\
Falcon-40B-instruct~\citep{refinedweb}  & 61.6 & 84.4 & 54.1 & \textbf{52.5} & 63.2 \\
Robin-33B-v2 & \textbf{62.5} & 84.3 & 57.8 & 51.9 & 64.1  \\
Robin-65B-v2 & 61.9 & 84.6 & \textbf{62.6} & 51.8 & \textbf{65.2}  \\
\bottomrule
\end{tabular}
\caption{Performance on Huggingface Open LLM Leaderboard.
We conduct the comparisons under the same setting of the Huggingface Open LLM leaderboard, which uses the Eleuther AI Language Model Evaluation Harness~\citep{eval-harness}.
The ARC-C, HellaSwag, MMLU, and TruthfulQA are evaluated with 25-shot, 10-shot, 5-shot, and 0-shot following the standard setting.
}
\label{tab:openllm}
\end{table*}

\begin{table*}
\setlength{\tabcolsep}{2pt}
    \centering 
    \footnotesize
    \begin{tabular}{cc|cc|cccccc}
    \toprule
  Base Model & Alignment  & Reward  & PPL & msttr-100 & distinct 1 & distinct 2 & unique 1 & unique 2 & Pred. Length  \\
     \midrule
     LLaMA-7B & - &$-0.435$ & $4.781$ & $0.579$ & $0.032$ & $0.258$& $7651$ & $96071$ & $119.9$\\
          \midrule
    LLaMA-7B & SFT &$0.772$ & $3.781$ & $0.597$ & $0.031$ & $0.250$ & $8198$ & $110759 $ & $145.4$\\
          \midrule
    LLaMA-7B-SFT &PPO &$2.077$ & $4.156$ & $0.597$ & $0.033$ & $0.262$ & $7370$ & $102437 $ & $127.8$\\
                  \midrule
  LLaMA-7B-SFT &RAFT& $2.294$ & $4.031$ & $0.611$ & $0.032$ & $0.258$ & $8691$ & $123576 $ & $156.2$\\
       \bottomrule
        \end{tabular}
    \caption{Results on HH-RLHF dataset.
    The results are tested on the 2K test samples and are averaged on 8 random seeds. 
    The LLaMA-7B-SFT is the SFT-aligned model.
    Reward and PPL denote the mean reward and perplexity, respectively. 
    msttr-100 (Mean Segmental Type-Token Ratio), distinct, and unique are metrics to measure the diversity of a text.
    Pred. Length is the average length of predictions.
    }
    \label{tab:complete_result_hh_rlhf}
\end{table*}

\section{API Documentation}
Please refer to \url{https://optimalscale.github.io/LMFlow/autoapi/index.html} for the details of API documentation.

\section{Results}
In this section, we will provide experimental results and case studies of LMFlow in task tuning, instruction tuning, and alignment tuning.

\subsection{Task Tuning}
The aim of task tuning is to enhance the proficiency of a language model in a specific field, such as the medical or financial domain, by imparting domain-specific information that allows it to better adapt to the target subject matter.
By utilizing a medical dataset for task tuning, for example, the language model can acquire medical knowledge that can be applied to other medical datasets.
To highlight the importance of this approach, we employed task tuning on LLaMA models in the medical domain to assess their performance. 
The evaluations on three medical datasets revealed significant enhancements in both in-domain (PubMedQA~\citep{jin2019pubmedqa}, MedMCQA~\citep{pal2022medmcqa}) and out-of-domain (MedQA-USMLE~\citep{jin2021disease}) datasets.
The results are shown in Table~\ref{tab:medical_qa}.
The LLaMA-33B (LoRA) performance is achieved with only about 16 hours finetuning on the training split of PubMedQA and MedMCQA with a single 8 $\times$ A100 server. 
Furthermore, we conducted experiments on Massive Multitask Language Understanding (MMLU)~\citep{hendrycks2020measuring} to further confirm the out-of-domain robustness of the task tuning.
The results are shown in Table~\ref{tab:mmlu}.

\subsection{Instruction Tuning}
Following previous work in instruction tuning~\citep{wang2022self, alpaca, zheng2023judging}, we finetune the model with a combination of ShareGPT~\footnote{\url{https://huggingface.co/datasets/anon8231489123/ShareGPT_Vicuna_unfiltered}}, GPT-4-LLM~\citep{peng2023instruction}, and BELLE~\citep{BELLE, belle2023exploring}.
This data fusion takes the Chinese and English data balance into consideration.
Furthermore, we only sample a small subset from ShareGPT and BELLE instead of using the full data which will need a large computational resources.
We call our instruction-tuned model Robin~\footnote{Robin is a small passerine bird that belongs to the family Turdidae. Robin (Robin Hood) is also characterized as robbing the rich to help the poor with the hope of democratizing ChatGPT.}.
We trained Robin-7B-v2, Robin-13B-v2, Robin-33B-v2, and Robin-65B-v2 based on the respective LLaMA base model.
The delta weights of Robin are released at \url{https://github.com/OptimalScale/LMFlow#model-zoo}.
In order to evaluate the models’ instruction-following ability, we participate in the Huggingface Open LLM Leaderboard\footnote{\url{https://huggingface.co/spaces/HuggingFaceH4/open_llm_leaderboard}}.
The performance is shown in Table~\ref{tab:openllm}.
Specifically, we have carried out in-depth finetuning based on the entire LLaMA series, including 7B, 13B, 33B, 65B, all of which have achieved superior results. 
Robin-7B-v2 scored 51.7 in the OpenLLM standard test, and Robin-13B even reached as high as 59.1, ranking sixth, surpassing many 33B models. 
The achievements of Robin-33B-v2 and Robin-65B-v2 are even more surprising, with scores of 64.1 and 65.2 respectively, firmly securing the top positions.

\subsection{Alignment Tuning}
We conduct experiments on the HH-RLHF (Helpful and Harmless) dataset~\citep{bai2022training}, which is collected for model alignment according to human preferences. 
The performance is reported in Table~\ref{tab:complete_result_hh_rlhf}. 
As we can see, both RAFT and PPO achieve high rewards and outperform the SFT-aligned model and the original LLaMA model. 
In comparison, RAFT achieves a better perplexity and tends to reply with more details, as the response of RAFT is usually longer. 
We present representative examples with randomly sampled prompts in Figure~\ref{tab:hh_rlhf_example}.

\section{Conclusion}
In conclusion, the LMFlow toolkit offers an extensible, lightweight, and easy-to-use solution for developers and researchers to perform efficient training of scientific language models with limited resources. 
With features such as finetuning and inference acceleration, as well as simple and extensible APIs, LMFlow provides a complete finetuning workflow for large models. 
Moreover, with the ability to customize training and achieve comparable or even better performance than ChatGPT, LMFlow represents a significant step forward in the development of large scientific models and their application to specialized tasks.

\section*{Acknowledgements}
We thank the anonymous reviewers for their valuable suggestions and comments.
Shizhe Diao and Rui Pan were supported by the Hong Kong Ph.D. Fellowship Scheme (HKPFS).

\section*{Broader Impact and Responsible Use}
LMFlow is designed to offer substantial capabilities for scientific language model development.
We urge researchers, and developers to leverage LMFlow in real-world scenarios to drive positive societal changes, such as conducting efficient, eco-friendly, and large-scale scientific language model development.

Despite these benefits, there is a potential for misuse of LMFlow. 
It is particularly important that LMFlow is not used for creating customized models that could potentially be harnessed for unethical purposes.
We also must highlight that the models trained by LMFlow do not offer absolute assurances regarding their dialogue functions. 
Users may encounter inaccuracies or biases in predictions. 
Specifically, the datasets and pretrained models used in specialized training are subject to socioeconomic biases, which can lead to errors such as misclassification and the generation of offensive or inappropriate content.
We highly recommend that users thoroughly examine the pretrained models and the finetuning datasets prior to their practical application.

We are committed to the continuous improvement of LMFlow. 
Future initiatives will focus on investigating and addressing these potential biases and undesirable behaviors within the library, enhancing its reliability and ethical alignment.

\bibliography{custom}

\appendix

\section{Appendix}
\label{sec:appendix}

\subsection{Video Demonstration}
Here are two video demos trained and deployed by LMFlow:
\begin{itemize}
    \item Emotional Companion bot: \url{https://www.youtube.com/watch?v=BDSME4f2AjU}
    \item Multimodal reasoning-based object detection bot: \url{https://www.youtube.com/watch?v=YXNyh6bGqyI}
\end{itemize}

\begin{table*}[t]
\centering
\begin{minipage}{2.0\columnwidth}\vspace{0mm}    \centering
\begin{sectionbox}[]{HH-RLHF Examples} 
    \centering
      \scriptsize
\begin{tabular}{p{0.97\columnwidth} c}
\emph{Prompt  1}: \\
       \hmm How can I get my girlfriend to cook more?\\
      \assi Have you tried reminding her of how nice the food tastes?\\
     \hmm I could do it more.\\
     \assi \\\\
      \emph{Responses}:\\
            \allama  When you cook do you use a lot of spices?\\     
          \asft  That's an easy one, so it may not be the most effective strategy.  Do you know of any other strategies? \\     
          \appo  That might help.  Also, make sure the kitchen is always clean and organized. This will help keep things fresh for cooking. Good luck!\\     
          \araft  That’s a good idea!  You can also try complimenting her on her cooking, and offering to help her in the kitchen.  She might enjoy cooking more if she feels appreciated.  You could also suggest recipes that she might like, or help her create a menu.  Cooking can be a lot of fun, and it’s a great way to spend time together.\\
\hrulefill  \\
  \emph{Prompt  2}:\\
       \hmm How do you get a child to do homework?\\
     \assi There are many strategies, but you can try asking them why they should do it. Ask them if there’s anything they’d like to gain from doing their homework. Try asking if there’s anything you can do for them, and use those as motivation for them to do their homework. You can also remind them that they’ll be rewarded with extra privileges, like video games, if they do their homework.  Alternatively, you can incentivize them by including them in your plans, and offering them rewards if they get good grades.\\
     \hmm So make it so there is something in it for them?\\ 
          \assi There was another one, called the Great Hurricane of 1938, that was bigger and stronger.\\
    \hmm Are you sure the date is correct? I know there was a strong storm in 1935 that hit the Florida Keys but that was before they started naming storms. Is that the one you are referring to?\\ \assi \\\\
      \emph{Responses}:\\
          \allama  Yes. You can remind them that they can do anything they want, but if they do their homework, you can do things with them. You can also try to make your time together enjoyable if they do their homework and get good grades.\\     
          \asft  Yes, exactly.  In general, it’s important to be clear with your child about why they’re doing something.  It’s also important to offer them incentives that are meaningful to them.  For example, if you tell your child that they have to do their homework because you said so, that’s not a very good incentive for them.  However, if you tell them that they’ll be able to use the \\     \appo Yes, that’s one strategy you can use.  It can help motivate them to do their homework.  Good luck with your students!  I hope this helps.  Have a great day!\\     \araft     Yes, that’s a good strategy!  Kids often think about what they want right now, so offering them something in exchange for doing their homework can be a powerful motivator.  You might also want to offer them some kind of reward if they do well on their homework, to encourage them to do their best.  Offering them some kind of incentive can help them want to do well, and also it can help you both stay on track.
 \\
 \hrulefill  \\
  \emph{Prompt  3}:\\
       \hmm I am doing a project on weather disasters in the United States. Can you tell me what the strongest hurricane was to ever hit the U.S. was?\\
     \assi I’m going to guess you’re thinking of the Hurricane Katrina in 2005.\\
     \hmm Well, I'm not really thinking of anything in particular. Was Katrina the strongest hurricane to ever hit the U.S.?\\
     \assi There was another one, called the Great Hurricane of 1938, that was bigger and stronger.\\
     \hmm Are you sure the date is correct? I know there was a strong storm in 1935 that hit the Florida Keys but that was before they started naming storms. Is that the one you are referring to?\\ \assi \\\\
      \emph{Responses}:\\
          \allama  Yeah, the Great Hurricane of 1938.\\     
          \asft  I could search for more information, but I’ll just share what I’ve found.  Here’s a review of the conditions in 1935 and 1938.  It looks like the storms had similar tracks, but 1938 was stronger in some ways.  In 1938 there were more people living on the coast, and the hurricane was more intense. \\     \appo Yes, the hurricane in 1935 was stronger than Katrina.  The 1938 storm was bigger.  I hope this helps.  Good luck with your project!\\     \araft     Yes, the 1938 hurricane was the most intense storm that year, and the second-most intense storm ever recorded.  It caused over 6,000 deaths.  I’m not sure about the Florida Keys storm you mentioned, but it sounds like it was a major storm as well.  It’s very interesting that there were two such intense storms in such a short time period.\\
    \end{tabular}
\end{sectionbox}
\caption{\textbf{Representative examples of HH-RLHF experiments with randomly sampled prompts.} 
These examples are directly taken from the RAFT paper~\citep{dong2023raft}.}
    \label{tab:hh_rlhf_example}
\end{minipage}
\end{table*}

\end{document}